\documentclass[letterpaper, 10 pt, conference]{ieeeconf}  

\IEEEoverridecommandlockouts                              

\pdfoutput=1

\usepackage[backend=biber,
            hyperref=true,
            url=false,
            isbn=false,
            doi=false,
            backref=false,
            style=ieee,
            natbib=true,
            mincitenames=1,
            maxcitenames=1,
            citestyle=numeric-comp,
            sorting=nyt,
            block=none]{biblatex}
        
\renewcommand{\bibfont}{\small}
\usepackage{graphics}
\usepackage{bbm}
\usepackage[pdftex]{graphicx}
\usepackage{wrapfig}
\DeclareGraphicsExtensions{.pdf,.png,.jpg}
\usepackage{epsfig}
\usepackage[font={small}]{caption}
\usepackage{subcaption}
\usepackage[rightcaption]{sidecap}
\usepackage{pbox}

\usepackage{bigstrut}
\usepackage{mathtools}
\usepackage{amsmath, amssymb, amscd}
\usepackage{ wasysym } 
\usepackage{amsfonts}
\usepackage{mathptmx} 
\usepackage{gensymb} 
\usepackage{nicefrac}       
\numberwithin{equation}{section} 

\DeclareMathAlphabet{\mathcal}{OMS}{lmsy}{m}{n}
\DeclareSymbolFont{largesymbols}{OMX}{cmex}{m}{n}
\usepackage{textcomp} 

\usepackage{algorithm} 
\usepackage{algorithmicx, algpseudocode} 

\usepackage{array} 
\usepackage{tabularx}
\usepackage{multirow}
\usepackage{multicol}
\usepackage{booktabs}
\usepackage{tabulary}

\usepackage[T1]{fontenc} 
\usepackage[utf8]{inputenc}
\usepackage[english]{babel} 
\usepackage{units}
\usepackage{bm}
\usepackage{times} 
\usepackage{xspace}
\usepackage{flushend}
\usepackage{balance} 
\usepackage{csquotes}
\usepackage{makeidx}
\usepackage{blindtext}

\let\labelindent\relax
\usepackage{enumitem}

\usepackage{ragged2e}
\usepackage{soul} 
\usepackage{subfiles} 

\usepackage[protrusion=true,expansion=true]{microtype}
\usepackage[yyyymmdd]{datetime}

\date{\protect\formatdate{1}{1}{2001}}

\usepackage{url}
\makeatletter
\g@addto@macro{\UrlBreaks}{\UrlOrds}
\makeatother
\usepackage{color}
\usepackage[usenames,dvipsnames,table,xcdraw]{xcolor}
\usepackage{pgfplots} 
\pgfplotsset{compat=newest}

\usepackage{marginnote}
\usepackage{soul} 

\usepackage[colorinlistoftodos]{todonotes}
\newcommand{\tocite}[1]{%
\textcolor{red}{[cite:\ifthenelse{\equal{#1}{}}{}{#1}?]}
}

\newcommand{\ignore}[1]{}
\newcommand{\etal}{et al.~}

\newcommand{\figref}[1]{Figure\,\ref{fig:#1}}

\newcommand{\seclabel}[1]{\label{sec:#1}}

\renewcommand{\baselinestretch}{1.0}



\newcommand{\states}{\mathcal{S}}
\newcommand{\actions}{\mathcal{A}}

\newcommand{\transitions}{\ensuremath{\mathcal{T}}}
\newcommand{\rewards}{\ensuremath{\mathcal{R}}}
\newcommand{\observations}{\ensuremath{\mathcal{Y}}}

\usepackage[colorinlistoftodos]{todonotes}

\title{\LARGE \bf
Mechanical Search: Multi-Step Retrieval \\ of a Target Object Occluded by Clutter%
}

\author{Michael Danielczuk$^{*1}$, Andrey Kurenkov$^{*2}$, Ashwin Balakrishna$^{1}$, Matthew Matl$^{1}$,\\David Wang$^{1}$, Roberto Mart\'{i}n-Mart\'{i}n$^{2}$, Animesh Garg$^{2}$, Silvio Savarese$^{2}$, Ken Goldberg$^{1}$%
\thanks{$^{*}$ Authors have contributed equally and names are in alphabetical order.}%
\thanks{$^{1}$University of California, Berkeley $^{2}$Stanford University}
}

\begin{document}

\maketitle

\begin{abstract}

When operating in unstructured environments such as warehouses, homes, and retail centers, robots are frequently required to interactively search for and retrieve specific objects from cluttered bins, shelves, or tables. \emph{Mechanical Search} describes the class of tasks where the goal is to locate and extract a known target object. In this paper, we formalize Mechanical Search and study a version where distractor objects are heaped over the target object in a bin. The robot uses an RGBD perception system and control policies to iteratively select, parameterize, and perform one of 3 actions -- push, suction, grasp -- until the target object is extracted, or either a time limit is exceeded, or no high confidence push or grasp is available. We present a study of 5 algorithmic policies for mechanical search, with 15,000 simulated trials and 300 physical trials for heaps ranging from 10 to 20 objects. Results suggest that success can be achieved in this long-horizon task with algorithmic policies in over 95\% of instances and that the number of actions required scales approximately linearly with the size of the heap. Code and supplementary material can be found at \url{http://ai.stanford.edu/mech-search}.
\end{abstract}

\section{Introduction}
\label{sec:introduction}

\begin{figure}[t!]
    \centering
    \includegraphics[width=0.95\linewidth]{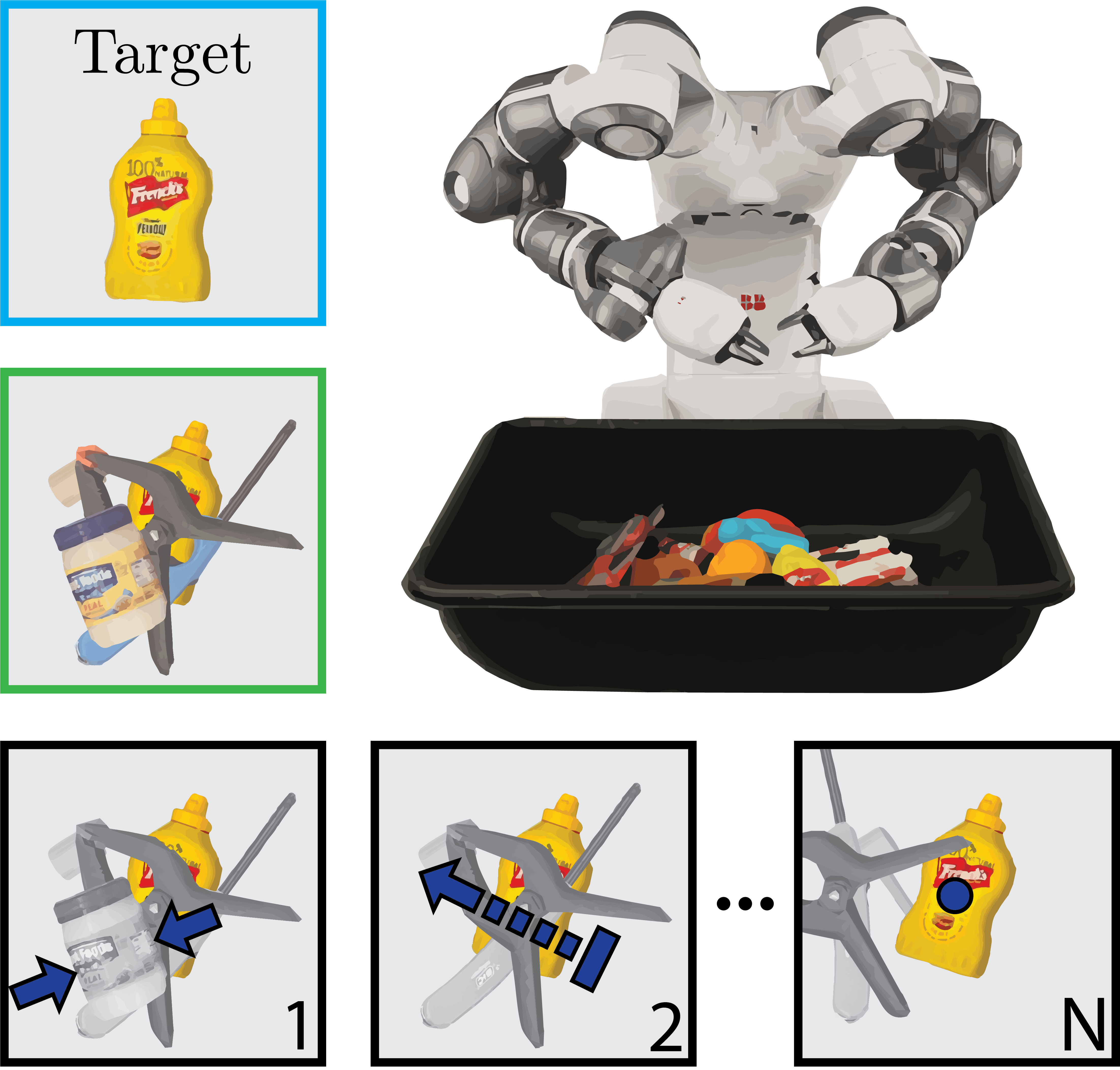}
    \caption{To locate and extract the target object from the bin, the system selects between 1) grasping objects with a parallel-jaw gripper, 2) pushing objects, or 3) grasping objects with a suction-cup gripper until the target object is extracted, a time limit is exceeded, or no high-confidence push or grasp is available.}
    \label{fig:intro}
\end{figure}

In unstructured settings such as warehouses or homes, robotic manipulation tasks are often complicated by the presence of dense clutter that obscures desired objects.
Whether a robot is trying to retrieve a can of soda from a stuffed refrigerator or pick a customer's order from warehouse shelves, the target object is often either not immediately visible or not easily accessible for the robot to grasp.
In these situations, the robot must interact with the environment to localize the target object and manipulate the environment to expose and plan grasps (see Figure \ref{fig:intro}). \emph{Mechanical Search} describes a class of tasks where the goal is to locate and extract the target object, and poses challenges in visual reasoning, task, motion, and grasp planning, and action execution.

Significant progress has been made in recent years on sub-problems relevant to Mechanical Search.
Deep-learning methods for segmenting and recognizing objects in images have demonstrated excellent performance in challenging domains~\cite{he2017mask,michaelis2018one,shaban2017one} and new grasp planning methods have leveraged convolutional neural networks (CNNs) to plan and execute high-quality grasps directly from sensor data~\cite{grasping1,grasping7,mahlerdexnet2}.
By combining object segmentation and recognition methods with action selectors that can effectively choose between different motion primitives in long horizon sequential tasks, multi-step policies can search for a target object and extract it from clutter.

In this paper, we propose a framework that integrates perception, action selection, and manipulation policies to address a version of the Mechanical Search problem, with 3 contributions:
\begin{enumerate}
\item A generalized formulation of the family of \textit{Mechanical Search} problems and a specific version for retrieving occluded target objects from cluttered bins using a series of parallel jaw grasps, suction grasps and pushes.
\item An implementation of this version, using depth-based object segmentation, single-shot image recognition, low-level grasp and push planners, and five action selection policies.
\item Data from simulation and physical experiments evaluating the performance of the five policies and that of a human supervisor. For simulated experiments, each policy was evaluated on a set of 1000 heaps of 10-20 objects sampled from 1600 3D object models; physical experiments used 50 heaps sampled from 75 common household objects.
\end{enumerate}
\section{Background and Related Work}
\seclabel{relwork}
\vspace{1mm}
\paragraph*{Perception for Sequential Interaction}
Searching for an object of interest in a static image is a central problem in active vision~\cite{van2011segmentation,redmon2016you,pmlr-v80-michaelis18a}. There has also been work on optimizing camera positioning for improving visual recognition (i.e., \emph{active perception}~\cite{5968,aydemir2011search}) and embodied interactions to explore (i.e., \emph{interactive perception}~\cite{bohg2017interactive,gupta2013interactive}). 
Mechanical Search differs from prior works in  interactive perception in that it deals with long grasping sequences.

Recent deep learning based methods achieve remarkable success in segmentation of RGB~\cite{ren2015faster,pinheiro2016learning} and depth images~\cite{chen20183d}, as well as in localizing visual templates in uncluttered~\cite{koch2015siamese,vinyals2016matching} and cluttered scenes~\cite{shaban2017one,michaelis2018one}. Furthermore, one-shot learning approaches using Siamese Networks for matching a novel visual template in images~\cite{koch2015siamese,vinyals2016matching} can translate well to pattern recognition in clutter~\cite{shaban2017one,michaelis2018one}.
We build on Mask R-CNN~\cite{he2017mask} by training a variant for depth-image based instance segmentation and leverage a Siamese network for target template matching for localization. 

\vspace{1mm}
\paragraph*{Grasping and Manipulation in Clutter}
Past approaches to this problem can be broadly characterized as model-based with geometric knowledge of the environment~\cite{berenson2008grasp,srivastava2014combined,moll2018randomized} and model-free with only raw visual input~\cite{saxena2008learning,katz2014perceiving,mahler2017learning}.
Recent studies have leveraged CNNs for casting grasping as a supervised learning problem with impressive results~\cite{grasping1,mahlerdexnet2,grasping3,Jang2017EndtoEnd,viereck2017learning,fang2018learning,kalashnikov2018qt}.
Pushing and singulation can facilitate grasping in cluttered scenes~\cite{chang2012interactive,hermans2012guided,danielczuklinear18}. Techniques for grasping in clutter, either as open-loop prediction or as closed-loop continuous control, have been studied but have not dealt with the multi-step plans that are critical to attain successful grasps on occluded or inaccessible target objects~\cite{kalashnikov2018qt,morrison2018closing}.
In contrast, we formulate Mechanical Search as an interactive search problem in significant clutter, necessitating a multi-step process combining grasping and pushing actions.

\vspace{1mm}
\paragraph*{Sequential Decision Making}
Sequential composition of primitives to enable long-term environment interaction has often been approached through hierarchical decomposition of control policies to manage task complexity. The idea of using hierarchical models for complex tasks has been widely explored in both reinforcement learning and robotics\,\cite{sutton1999between, kober2013reinforcement,sung2013learning}. 
Training such multi-level models can be computationally expensive and has been limited to either simulated or elementary physical tasks~\cite{duan2017one, xu2017neural}.

\vspace{1mm}
\paragraph*{Search Based Methods}
Traditional task planning approaches abstract away perception and focus on high-level task plans and low-level state spaces~\cite{fikes1971strips,srivastava2014combined}.
For instance, in robotic applications, hierarchical methods have been used to learn task planning strategies while abstracting away low-level motion planning~\cite{paxton2016want,srivastava2014combined,wolfe2010combined}. However, high-level planning requires complete domain specification a priori, and complex geometric and free space reasoning make this approach applicable only to uncluttered environments with few objects, such as a tabletop with one or two objects.

A similar problem has been studied in the context of mobility under problem domains of target-driven and semantic visual navigation~\cite{zhu2017target, mousavian2018visual, gupta2017cognitive}. These studies look at finding visual targets in unknown environments without maps through sensory pattern matching. 
The work by Gupta \etal~\cite{gupta2013interactive} is the closest to the approach considered in this paper. Their work also considers the problem of searching for a specific object using pushing and grasping actions, but when the objects are arranged in a shelf. 
We consider significantly more cluttered settings, while also executing temporally extended manipulation policies.
\section{Mechanical Search: Problem Formulation}
\label{sec:prob_form}
\begin{figure*}[t!]
\vspace{8pt}
\centering
\includegraphics[width=0.95\linewidth]{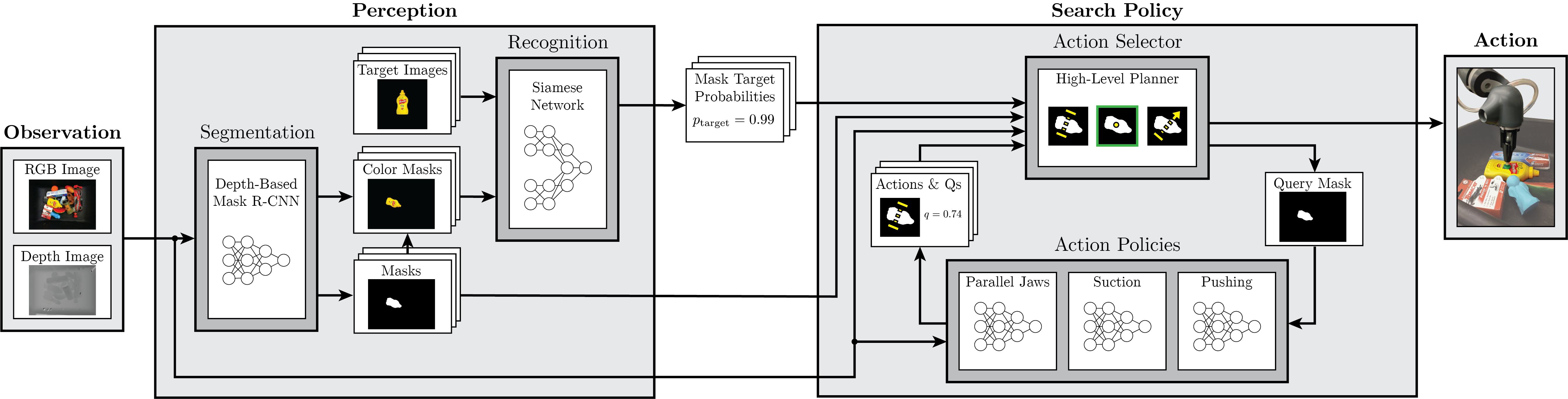}
\caption{System architecture. At each timestep, the RGB-D image of the bin is segmented using a variant of Mask R-CNN trained on synthetic depth images. The colorized masks are each assigned a probability of belonging to the target object using a Siamese Network as a pattern comparator. These masks are then fed to an action selector, which chooses which object to manipulate and passes that object mask as context to all the action policies. These policies each compute an action with an associated quality score and pass them back to the action selector, which then chooses an action and executes it on the physical system. This process continues until the target object is retrieved. In simulation, the perception pipeline is removed and planners operate on full-state information rather than object masks.}
\label{fig:system}
\end{figure*}

In Mechanical Search, the objective is to retrieve a specific target object ($x^{*}$) from a physical environment ($E$) containing a variety of objects $X$ within task horizon $H$ while minimizing time. The agent is initially provided with a specification of the target object in the form of images, text description, a 3D model, or other representation(s). We can frame the general problem of Mechanical Search as a Partially Observable Markov Decision Process (POMDP), defined by the tuple $(\states, \actions, \transitions, \rewards, \observations)$.

\begin{itemize}

\item {\bf States ($\states$)}. A bounded environment $E$ at time $t$ containing $N$ objects $\mathbf{s}_t = \lbrace \mathcal{O}_{1,t}, ..., \mathcal{O}_{N,t}\rbrace$. Each object state $\mathcal{O}_{i,t}$ includes a ground truth triangular mesh defining the object geometry and pose. Each state also contains the pose and joint states of the robot as well as the poses of the sensor(s). 
\item {\bf Actions ($\actions$)}. A fixed set of parameterized motion primitives.
\item  {\bf Transitions ($\transitions$)}. Unknown transition probability distribution $P: \states \times \states \times \actions \rightarrow \left[0, \ \infty \right)$.
\item {\bf Rewards ($\rewards$)}. Function given by $R_t = R(\mathbf{s}_t, \mathbf{a}_t) \rightarrow \mathbb{R}$ at time $t$ that estimates the change in probability of successfully extracting the target object $x^{*} \in X$ within task horizon $H$.
\item {\bf Observations ($\observations$)}. Sensor data, such as an RGB-D image, $y_t$ from robot's sensor(s) at time $t$ (see Fig.~\ref{fig:intro}).
\end{itemize}

In this paper, we focus on a specific version of Mechanical Search: extracting a target object specified by a set of $k$ RGB images from a heap of objects in a single bin while minimizing the number of actions needed. For this problem, we precisely specify the observations, the action set, and the reward function. All other aspects of the problem formulation are sufficiently captured by the general POMDP formulation above.

\begin{itemize}
\item {\bf Observations}. An RGB-D image from an overhead camera.
\item {\bf Actions.}
    
\begin{itemize}
[   topsep=0pt,
    noitemsep,
    leftmargin=2pt,
    itemindent=12pt]
    \item \textbf{Parallel Jaw Grasping}: A center point $\mathbf{p} = (x,y,z) \in \mathbb{R}^3$ between the jaws, and an angle in the plane of the table $\phi \in \mathbb{S}^1$ representing the grasp axis~\cite{mahlerdexnet2}. 
    
    \item \textbf{Suction Grasping}: A target point $\mathbf{p} = (x,y,z) \in \mathbb{R}^3$ and spherical coordinates $(\phi,\theta) \in \mathbb{S}^2$ representing the axis of approach of the suction cup~\cite{mahler2017dex}.
    
    \item \textbf{Pushing}: A linear motion of the robot end-effector between two points $\mathbf{p}$ and $\mathbf{p}'\in \mathbb{R}^3$. 
    
\end{itemize}

\item {\bf Reward.} Let $v_t$, derived from $y_t$, denote the estimated grasp reliability on the target object. An intuitive reward function would be the increase in estimated grasp reliability on the target object:
$$R(s_t, a_t) = v_{t+1} - v_{t}$$
The policies used in this paper do not directly optimize this reward function because it is difficult to compute; instead, they continue to remove and push objects via heuristic methods until the target object is extracted. In future work, we will develop methods to approximate this function.
\end{itemize}

\section{Perception and Decision System}
\label{sec:system_overview}
As shown in Figure~\ref{fig:system}, we implement the system both in simulation and for physical experiments via a pipeline for perception and policy execution.

\subsection{Perception}
The system first processes the RGB-D image into a set of segmentation masks using an object instance segmentation pipeline trained on synthetic depth images. Then, a Siamese network is used to attempt to identify one of the masks as the target object, and a target mask is returned if a high confidence match is found. If no high confidence match is found, the perception system reports that no masks match the target object.

\paragraph*{Object Instance Segmentation} 
We first compute a mask for each object instance. Each mask is a binary image with the same dimensions as the input RGB-D image. These masks are computed with SD Mask R-CNN, a variant of Mask R-CNN trained exclusively on synthetic depth images~\cite{danielczuk2018segmenting}. It converts a depth image into a list of unclassified binary object masks, and generalizes well to arbitrary objects without retraining. Recent results suggest that depth cues alone may be sufficient for high-performance segmentation, and this network's generalization capabilities are beneficial in a scenario where only the target object is known and many unknown objects may be present.

\paragraph*{Target Recognition}
Next, the set of masks is combined with the RGB image to create color masks of each object. Each of the $m$ color masks is cropped, scaled, rotated, and compared to each of the $k$ images in the target object image set using a Siamese network~\cite{koch2015siamese}. For each pair of inputs, the Siamese network outputs a recognition confidence value between 0 and 1, with a mask's recognition confidence score set to the maximum recognition confidence value over the $k$ target object images. If the mask with the highest score has a score above recognition confidence threshold $t_{r}$, the mask is labeled as the target object. Otherwise, we report that no masks match the target object. See the appendix for training and implementation details. 

\subsection{Search Policy}
\label{subsec:action_block}

Given the RGB-D image and the output of the perception pipeline, the system executes the next action in the search procedure by selecting the object to act on and the action to perform on it. Our approach to the version of Mechanical Search described in Section~\ref{sec:prob_form} for bin picking includes searching for actions in three continuous spaces (parallel jaw grasp, suction grasp and push). However, more complex versions of Mechanical Search (e.g., search for an object in a house) could have even more complex search spaces (e.g., navigation). To allow our method to scale to these more complex versions, we propose a hierarchical approach: (1) an action selector that queries a set of action policies on a specific object for a particular action and associated quality metric and (2) action policies that correspond to the possible actions in the problem formulation. 
 
\paragraph* {Action Selection} 
The search policy first determines which object masks to send to the action policies. Then, using the actions and associated quality metric returned by the low level policies, the high level planner determines whether to execute the action in the environment.

The action selector takes as input from the perception system the set of all $m$ visible object masks ($[o_1, \hdots o_m]$), possibly including an object mask that is positively identified as the target object ($o_T$), from the perception system. It then selects an action policy and a goal object, $o_{goal}$, from $[o_1, \hdots o_m]$ and sends the action policy a query $q(o_{goal})$. The action policy $p_i$ responds with an action $a_i = p_i(o_{goal})$ and a quality metric $Q(a_i, o_{goal})$ for the action, which is used to decide whether to execute the action.

\paragraph* {Action Policies} 
Each action policy $p_i$ takes as input an object mask from the action selector ($o_{goal}$) and the RGBD image observation and returns an action $a_i = p_i(o_{goal})$ and a quality metric $Q(a_i, o_{goal})$. In simulation, the object masks and depth images are generated from ground-truth renderings of each object, while in physical experiments, depth images are obtained using a depth sensor and object masks are generated by the perception pipeline. The set of action policies in our system are:

\paragraph*{Parallel Jaw Grasping} 
In simulation, pre-computed grasps are indexed from a Dex-Net 1.0 parallel-jaw grasping policy~\cite{mahler2016dex}, and the grasp with the highest predicted quality on $o_{goal}$ is returned as the action along with an associated quality metric. For physical experiments, parallel-jaw grasps are planned using a Dex-Net 2.0 Grasp Quality CNN (GQ-CNN)~\cite{mahlerdexnet2}. To plan grasps for a single object in a depth image, grasp candidate sampling is constrained to the goal object's segmentation mask. The GQ-CNN evaluates each candidate grasp and returns the grasp with the highest predicted quality and its associated quality metric.

\paragraph*{Suction Grasping}
For simulation experiments, grasp planning is done with a Dex-Net 1.0 suction grasping policy~\cite{mahler2016dex}. For physical experiments, suction cup grasps are planned with a Dex-Net 3.0 GQ-CNN~\cite{mahler2017dex}, with mask-based constraints to plan grasps only on the goal object's segmentation mask. The GQ-CNN evaluates each candidate grasp and returns the grasp with the highest predicted quality and its associated quality metric.

\paragraph*{Pushing} 
The pushing action policy, similar to that in~\cite{danielczuklinear18}, selects $\mathbf{p}'$ as the most free point in the bin. This point is computed by taking the signed distance transform of a binary mask of the bin walls and objects, finding the pixel with the maximum signed distance value, and deprojecting that pixel back into $\mathbb{R}^3$. Given an object to push, $\mathbf{p}$ is then selected so that the gripper is not in collision at $\mathbf{p}$, the line from $\mathbf{p}$ to $\mathbf{p}'$ passes through the object's center of mass, and the push direction is as close as possible to the direction of the most free point in the bin. The pushing policy returns the push satisfying the above constraints as its action if one exists. The returned quality metric is $1$ if a valid push exists and $0$ if not.
\section{Action Selection Policies}
All action selection methods use input from the perception system to generate a specific object priority list. Each action selection method generates a priority list in a different way but all have the same action execution criteria. For all action selection methods described here, a grasp action is executed if the quality metric returned by the action policy exceeds $t(o)$, the grasp confidence threshold for object mask $o$. The grasp confidence threshold for the object mask positively identified as the target object $o_T$ is given by $t(o_T) = t_{\text{thresh}}$. For policies without pushing, $t(o) = t_{\text{thresh}}, \forall o$, while for policies with pushing, $t(o) = t_{\text{high}}, \forall o \neq o_T$. Policies with pushing can be more conservative in their choice of grasps, so they use a higher grasp confidence threshold $t_{\text{high}}$ for non-target objects. A push action is performed if a valid push is found (quality 1). Details on parameters used can be found in the appendix.

Each action selection method iterates through its priority list, queries the grasping action policies for each object mask, and executes the returned action with the highest quality metric among the two grasping policies if it satisfies the action execution criteria. If the target object is grasped, the policy terminates and reports a success. If no grasping action satisfies the criteria and the policy does not have pushing, the policy terminates and reports a failure. If the policy does have pushing, it iterates through its priority list, queries the pushing action policy for each object mask, and executes the first action that satisfies the criteria. If no pushing action satisfies the criteria, or if a pushing action has been selected more than three consecutive times, the policy terminates with a failure.

\paragraph*{Action Selection Methods}
The action selection methods are distinguished by whether or not they have pushing as an available action policy and by their generated object priority list:

\begin{enumerate}
    \item \textbf{Random Search:} Prioritizes objects randomly, with no preference for the target object mask ($o_T$).
    \item \textbf{Preempted Random Search (with and without pushing):} Always prioritizes $o_T$ and prioritizes other objects randomly.
    \item \textbf{Largest-First Search (with and without pushing):} Always prioritizes $o_T$ and ranks the other objects by their visible area. If the target object isn't visible, this strategy will increase the likelihood of removing objects that may be occluding the target object.
\end{enumerate}

\paragraph*{Termination Criteria}
In addition to the termination criteria outlined above (terminate and return success if target object grasped, return failure if no good grasp/valid push found), we impose two more termination conditions on our policies which cause them to return a failure: (1) $2N$ timesteps have elapsed, where $N$ is the initial number of objects in the bin and (2) The target object is inadvertently removed from the work space when another object is grasped or pushed.

\section{Experiments} 
\label{sec:experimental-setup}

\subsection{Simulation} \label{sec:sim-exp-procedure}

\paragraph*{Heap Generation}  Three datasets of simulated heaps are generated, each containing 1000 heaps of $N$ objects, for $N \in \{10, 15, 20\}$. Then, using the Bullet Physics Engine~\cite{coumans2017bullet}, sampled objects are dropped one by one into the bin, and the target object is chosen to be the most occluded object. Please refer to the appendix for further details.

\paragraph*{Rollouts} To simulate grasp actions, we use the same approach as in~\cite{mahler2017learning}: using wrench space analysis, we determine whether or not an object can be lifted from the heap ~\cite{prattichizzo2008grasping,ten2018using}. If the object can be lifted, a constant upward force is applied to the object's center of mass until it leaves the bin, and the remaining objects are allowed to come to rest. To simulate push actions, we check that the gripper can be placed in the starting location without collisions, and only execute pushes if this is the case. Then, we place a 3D model of the closed gripper in the physics simulator and move it from the start point to the end point of the push, as in~\cite{danielczuklinear18}. 

\begin{figure}[t!]
\vspace{8pt}
\centering
\includegraphics[width=\linewidth]{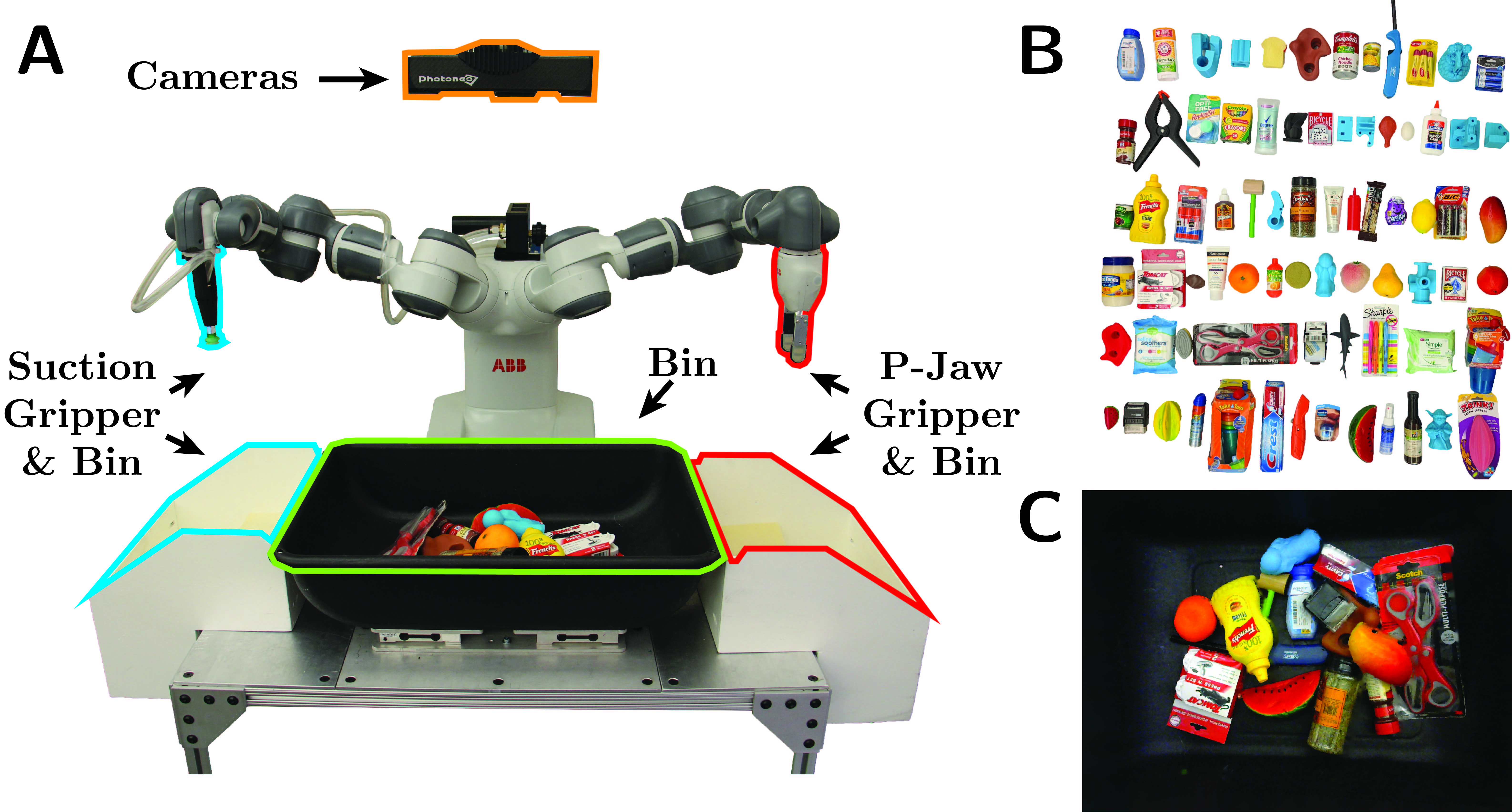}
\caption{\textbf{(A)} Front view of the robot and bin setup. The black bin is the primary bin in which heaps are initialized, and the white bins provide space for the robot to deposit grasped items. \textbf{(B)} The 75 objects used in physical experiments. \textbf{(C)} A sample heap of 15 objects used in the physical experiments.}
\label{fig:realheaps}
\end{figure}

\begin{figure*}[t!]
\vspace{4pt}
\centering
  \includegraphics[width=\linewidth]{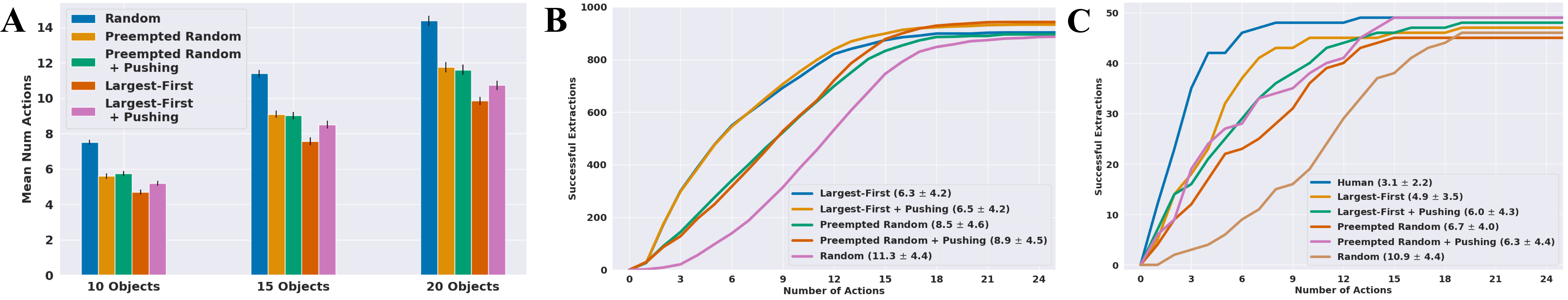}
    \caption{Performance of policies on \textbf{(A)} simulated heaps of 10, 15, and 20 objects over a total of 15,000 simulated rollouts and 300 physical rollouts, \textbf{(B)} simulated heaps of 15 objects, and \textbf{(C)} real heaps of 15 objects. The largest-first search policies are the most efficient, and are able to extract the target object in the least number of actions. All policies have similar reliability, although pushing shows potential to avoid more failures in simulation. The human was allowed to look at the RGBD image inputs and choose an object to push or grasp. Means and standard deviations for successful extractions are shown in parentheses for each policy.}
  \label{fig:results}
\end{figure*}

\subsection{Physical} \label{sec:phys-exp-procedure}

\paragraph*{Heap Generation}
We randomly sample 50 heaps of 15 items each from a set of 75 common household objects with relatively simple shapes, such as boxes and cylinders, as well as more complex geometries, such as plastic climbing holds and scissors (see Figure~\ref{fig:realheaps}). We also include several 3D-printed items, which present a challenge for both segmentation and target object recognition due to their unusual shapes and uniform texture. 
A target object is chosen at random from each 15 item heap. Then, in order to generate adversarial bin configurations, each rollout is initialized by first shaking the target object in a box to randomize its pose and dumping into the center of the bin, and then shaking the other fourteen objects and pouring them over the target object.

\paragraph*{Policy Rollouts} \label{sec:policy_rollouts} We execute pushing and grasping actions on an ABB YuMi robot equipped with suction-cup and parallel-jaw grippers (see Figure \ref{fig:realheaps}). Actions generated by the search policy are transformed into a sequence of poses for the robot's end-effectors, and we use ABB's RAPID linear motion planner and controller to execute these motions.

\paragraph*{Human Supervisor Rollouts} 

For comparison, we also benchmark a human supervisor's performance as an action selector. At each timestep, the human is asked to draw a mask in the scene on which to plan a push or a grasp. Then, grasps and pushes are planned and executed on the specified mask with the same action primitives described above (parallel jaw grasps, suction grasps, linear pushes). Thus, the human is limited by the available action primitives, but is allowed to use their own judgement for perceptual reasoning and high level action planning.

\subsection{Evaluation Metrics} We evaluate each policy according to its reliability and efficiency in target object retrieval. Reliability is defined as the frequency at which the target object is successfully extracted, while efficiency is defined as the mean number of actions taken to successfully extract the target object.
For each experiment, we recorded the number of successes and failures, as well as statistics regarding the number of actions taken.
\section{Results}

\subsection{Simulation Results}
We tested each action selection policy in simulation with heaps generated using the method described in \ref{sec:experimental-setup} and running until termination on each heap. A total of 15,000 simulation experiments were conducted over all policies. The results for each of the five policies are shown in \figref{results}, and a detailed breakdown of each policy can be found in the appendix. \figref{results}(A) shows the mean number of actions needed for each policy as a function of heap size. These results suggest that extracting the target object becomes more difficult as heap size increases and the mean number of actions needed for each policy appears to scale linearly with heap size, although the rate of increase is not constant across policies.
 
\figref{results}(B) shows cumulative successful extractions for a given amount of actions (number of successful extractions in that many actions or fewer). All policies have success rates of 90\% or higher on 15 object heaps. The cumulative success plot suggests that grasping the target object when possible provides improvement over the random policy, and that prioritizing larger object masks when the target object is inaccessible further increases efficiency.
The largest-first policies successfully extract the target object within 5 or fewer actions on 50\% of the heaps, while the preempted random and random policies only do so for 30\% and 10\% of heaps respectively. 

Results also suggest that pushing can increase overall success rate, as policies that included pushing succeeded on at least 3\% more heaps than those without pushing. For policies with pushing, only 5\% of all actions attempted are pushes, as opposed to 29\% parallel jaw grasps and 66\% suction grasps, on ten object heaps. The percentage of push actions decreases further when increasing the number of objects to just 3\% for 15 object heaps. We suspect that the reason pushes are selected so rarely is because the pushing primitive is designed to execute a sequence of linear pushes to singulate a particular object, rather than flatten a heap of objects. While the former directly addresses the objective of the push, the latter is much easier to achieve in practice, since in many cases, especially with many objects in the bin, it may be almost impossible to plan collision-free pushes that successfully singulate the object of interest.

In simulation, failures account for 6-12\% of all rollouts. These failures fall into three categories: 1) the policy fails to plan an action (e.g., no actions are available on any remaining objects with a quality metric above the threshold), 2) the target object is inadvertently removed from the bin, despite an attempted action that was not a grasp on the target object, or 3) the policy reaches the maximum number of timesteps given to extract the object ($2N$ timesteps, where $N$ is the initial number of objects in the heap). 

Failure mode (1) accounts for 85-90\% of all failures, while (2) accounts for nearly all of the remaining failures. Mode (1) typically happens because objects are moved to the corners of the bin, making it difficult to plan collision-free grasps. Mode (2) occurs more often for policies that include pushing actions, since pushing in clutter can occasionally lead to objects being pushed up and over the bin walls. It is also possible that the target object is removed from the work space when another object is grasped. Mode (3) never actually occurs in any of our experiments, and the maximum timesteps cutoff is intentionally set high to exhaust the policy of actions.

\subsection{Physical Results}
\figref{results}(C) shows results on the physical system. A total of 300 physical experiments were conducted over all policies. All policies retrieved the target object within the given number of timesteps at least 90\% of the time, and success rates were not statistically different between policies. However, the cumulative success curves suggest similar trends to those seen in the simulation results, with the largest-first policies outperforming the preempted random and random policies in terms of efficiency. The largest-first policies again successfully extract the target object within 5 or fewer actions on on 50\% of the heaps, while the  preempted random and random policies only do so for 40\% and 10\% of heaps respectively. 

Stochasticity in the initial bin state in physical experiments can result in varying difficulty for different policies on the same heap. For example, a target object may be completely covered by other objects when one policy is presented with a given heap, but for another policy, the target may be partially or fully visible in the initial state. Thus, policies may occasionally get ``lucky" or ``unlucky" with respect to the target object visibility in the initial state, which may account for some increased variance in the physical results.

Failure cases for the physical heaps are very similar to those in simulation: 93\% of failures arise from the policy being unable to plan an action. However, in physical experiments, out of the 300 heaps evaluated for all policies, only 1 rollout failed due to timing out. Failure to plan actions is almost always due to the target object lying flat on the bottom of the bin (e.g., the dice, sharpie pens, or another blister-pack object), making it difficult to obtain accurate segmentation. Another common reason for failure to plan actions is when no mask is identified as the target object, which often occurs for 3D printed objects.

\subsection{Action-Limited Human Supervisor}
The human supervisor outperforms all policies presented here, requiring an average of just 3.1 actions to extract the target object due to more intelligent action selection. Specifically, we noticed that a human operator chose to push far more frequently (26\% of all actions, compared to 6\% for the other action policies with pushing), especially when objects were heaped in the center of the bin and the target was not visible. These pushes tend to spread many objects out over the bottom of the bin, as opposed to a grasping action that would remove only a single object from the top of the heap.
\section{Discussion and Future Work}

We present a general formulation for mechanical search problems and describe a framework for solving the specific problem of extracting a target object from a cluttered bin.
While the best action selection method (largest-first) is much more efficient than random search and provides a solid baseline, a human selecting the low-level actions can still achieve 37\% higher efficiency by pushing significantly more effectively and often. We will explore how reinforcement learning in simulation can address this gap.

In future work, the action primitives used (grasp, suction, push) can also be improved. We conjecture that more effective push primitives can be learned from simulation. 

\section{Appendix}

\subsection{Extended Results}
Tables~\ref{tab:sim_results},~\ref{tab:sim_actions},~\ref{tab:phys_results}, and~\ref{tab:phys_actions} give a detailed breakdown of each policies selected actions and success rate over the 1000 simulated trials and 50 trials on the physical robot for each policy on 15 object heaps. In simulation, pushing actions result in higher success rates. On the physical system, the human policy selects pushing actions much more frequently to clear multiple occluding objects from the target object. We will explore this discrepancy further in future work.

\begin{table}[h!]
    \centering
    \begin{tabular}{@{} l *2c @{}}    \toprule
\emph{Simulation Policy} & \emph{Success Rate} & \emph{Mean Actions}  \\\midrule
Random    & 88.8\%  & $11.26 \pm 0.15$\\ 
Preempted Random & 89.7\% & $8.55 \pm 0.16$\\ 
Preempted Random + Pushing & 94.3\% & $8.87 \pm 0.15$\\
Largest-First & 90.3\% & $6.35 \pm 0.14$ \\
Largest-First + Pushing & 93.3\% & $6.51 \pm 0.14$\\\bottomrule
 \hline
\end{tabular}
    \caption{Success rate and mean number of actions (with standard error of the mean) for extraction for 1000 trials of each policy tested in simulation. The largest-first policies extract the target most efficiently, and pushing shows ability to increase overall success rate.}
    \label{tab:sim_results}
\end{table}

\begin{table}[h!]
    \centering
    \begin{tabular}{@{} l *3c @{}}    \toprule
\emph{Simulation Policy} & \emph{Suction} & \emph{Parallel-Jaw} & \emph{Push}\\\midrule
Random    & 7015 & 4467 & 0\\ 
Preempted Random & 6168 & 2870 & 0\\ 
Preempted Random + Pushing & 6180 & 2825 & 274 \\
Largest-First & 5266 & 1718 & 0\\
Largest-First + Pushing & 5116 & 1706 & 259 \\\bottomrule
 \hline
\end{tabular}
    \caption{Breakdown of action selection for each policy in simulation over 1000 trials. Policies typically attempt many more suction grasps due to better accessibility in clutter, and only attempt pushes a small fraction of the time.}
    \label{tab:sim_actions}
\end{table}

\begin{table}[h!]
    \centering
    \begin{tabular}{@{} l *2c @{}}    \toprule
\emph{Physical Policy} & \emph{Success Rate} & \emph{Mean Actions}  \\\midrule
Random    & 92\%  & $10.87 \pm 0.66$\\ 
Preempted Random & 90\% & $6.71 \pm 0.61$\\ 
Preempted Random + Pushing & 98\% & $6.31 \pm 0.63$\\
Largest-First & 94\% & $4.85 \pm 0.51$ \\
Largest-First + Pushing & 96\% & $6.00 \pm 0.63$\\
Human & 98\% & $3.06 \pm 0.32$ \\\bottomrule
 \hline
\end{tabular}
    \caption{Success rate and mean number of actions (with standard error of the mean) for extraction for 50 trials of each policy tested on the physical system. All policies achieve success rates of over 90\% due to effective low-level grasping policies, but the human outperforms the best policy by 37\% in terms of mean number of actions, suggesting that there is considerable room for action selection policy improvement.}
    \label{tab:phys_results}
\end{table}

\begin{table}[h!]
    \centering
    \begin{tabular}{@{} l *3c @{}}    \toprule
\emph{Physical Policy} & \emph{Suction} & \emph{Parallel-Jaw} & \emph{Push}\\\midrule
Random    & 275 & 300 & 0\\ 
Preempted Random & 282 & 76 & 0\\ 
Preempted Random + Pushing & 250 & 58 & 19 \\
Largest-First & 222 & 47 & 0\\
Largest-First + Pushing & 244 & 59 & 18 \\
Human & 99 & 27 & 44 \\\bottomrule
 \hline
\end{tabular}
    \caption{Breakdown of action selection for each policy in 50 physical trials. The human pushes much more frequently than the other policies, especially to clear multiple occluding objects at the beginning of the trial.}
    \label{tab:phys_actions}
\end{table}

\subsection{Siamese Network Implementation Details}
The Siamese network architecture involves first passing each input $512 \times 512$ RGB image through a ResNet-50 architecture pretrained on ImageNet. During training of the Siamese network, these weights remained fixed. The featurizations of the input images are then concatenated and passed through two dense, fully connected layers: the first with 1024 neurons and ReLU activations, and the second with a single output neuron and sigmoid activation, whose output represents the probability that the two input images are of the same object. The motivation for this architecture is to allow the Siamese network to learn a distance metric over the ResNet-50 featurizations. The training dataset for the Siamese network consists of 5 views of each of the objects used in physical experiments. For each view, we generated a total of 10 additional images: 5 randomly rotated versions of the original image as well as 5 rotated versions that are partially occluded. To simulate occlusions, we took randomly rotated and scaled binary masks of a dataset of synthetic objects, and overlay the masks on the original object. We only used occlusions that covered at least 20\% and at most 80\% of the original pixels of the object. For training, we sampled 10,000 positive and 10,000 negative image pairs, where a positive image pair consists of an original image of an object and one of the 10 augmented images and a negative image pair consists of an original image of an object and one of the 10 augmented images of an entirely different object. The network is then trained with a contrastive loss function for 10 epochs using a batch size of 64 and the Adam optimizer with a learning rate of 0.0001. For physical experiments, a recognition confidence threshold $t_{r} = 0.9$ was used.

\subsection{Simulated Heap Generation}
Simulated heaps are generated by sampling: 1) $N$ objects from a dataset of over 1600 3D models, 2) a heap center around the center of the bin, and 3) planar pose offsets for each object around the heap center. Then, using the Bullet Physics Engine, sampled objects are dropped one by one into the bin from a fixed height at their pose offset from the heap center, and all objects are allowed to come to rest (i.e. all velocities of all objects go to zero). Once all objects have been added to the heap, the modal and amodal segmentation masks for each object are rendered from the camera's perspective.  The modal segmask of an object is a segmask of the portion of the object visible from the perspective of the camera (accounting for occlusions), while the amodal segmask of an object is a segmask of the object's exact position in the scene given ground truth information from the simulation environment. Using these masks, the target object is chosen to be the object with the smallest ratio between modal and amodal segmask area (i.e., the least visible object in the bin). This metric is used as a proxy for finding the most occluded object.

\subsection{Policy Parameters}

\paragraph*{Simulation Policy Parameters} Grasp confidence thresholds of $t_{\text{thresh}} = 0.15$ and $t_{\text{high}} = 0.3$ are used in simulation for the high-level action selector to determine whether to execute grasp actions from the low level grasp policies. In experimental trials, these values were found to provide a balance between avoiding grasp failures and quickly clearing objects from the bin as soon as sufficiently good grasps become available.

\paragraph*{Physical Policy Parameters} In physical experiments, grasp confidence thresholds of $t_{\text{thresh}} = 0.1$ and $t_{\text{high}} = 0.3$ were used for action selection to determine whether to execute grasp action plans from the low-level grasp action policies. These values are similar to those used in simulation, but $t_{\text{thresh}}$ is made slightly lower for physical experiments since it we observed that low confidence grasps succeeded more often in physical experiments than in simulation, which was designed to be conservative to encourage good transfer to reality. Additionally, $t_{\text{high}} = 0.5$ was used for policies that included low-level pushing action policies, so that pushing would be further encouraged over low-quality grasp actions.

\section{Acknowledgments}
\footnotesize
This work is partially supported by a Google Focused Research Award and was performed jointly at the AUTOLAB at UC Berkeley and at the Stanford Vision \& Learning Lab, in affiliation with the Berkeley AI Research (BAIR) Lab, Berkeley Deep Drive (BDD), the Real-Time Intelligent Secure Execution (RISE) Lab, and the CITRIS "People and Robots" (CPAR) Initiative. Authors were also supported by the SAIL-Toyota Research initiative, the Scalable Collaborative Human-Robot Learning (SCHooL) Project, the NSF National Robotics Initiative Award 1734633, and in part by donations from Siemens, Google, Amazon Robotics, Toyota Research Institute, Autodesk, ABB, Knapp, Loccioni, Honda, Intel, Comcast, Cisco, Hewlett-Packard and by equipment grants from PhotoNeo, and NVidia. This article solely reflects the opinions and conclusions of its authors and do not reflect the views of the Sponsors or their associated entities. We thank our colleagues who provided helpful feedback, code, and suggestions, in particular Jeff Mahler.

\renewcommand*{\bibfont}{\footnotesize}
\printbibliography 

\end{document}